\documentclass{article}
\usepackage[final]{style/neurips_2024}
\usepackage{booktabs}
\usepackage{multirow}
\usepackage{multicol}
\usepackage{graphicx}
\usepackage{enumitem}
\usepackage{xspace}
\usepackage{wrapfig}
\usepackage{paralist}
\usepackage[T1]{fontenc}

\setitemize{noitemsep,topsep=0pt,parsep=0pt,partopsep=0pt}

\usepackage[dvipsnames,table,xcdraw]{xcolor}
\usepackage{circledsteps}
\usepackage{bbding}
\usepackage{makecell}
\usepackage{anyfontsize}
\usepackage{tcolorbox}
\usepackage{listings}
\definecolor{codegreen}{rgb}{0,0.6,0}
\definecolor{codegray}{rgb}{0.5,0.5,0.5}
\definecolor{codepurple}{rgb}{0.58,0,0.82}
\definecolor{backcolour}{rgb}{0.95,0.95,0.95}

\lstdefinestyle{mystyle}{
    backgroundcolor=\color{backcolour},   
    commentstyle=\color{codegreen},
    keywordstyle=\color{codepurple},
    numberstyle=\tiny\color{codegray},
    stringstyle=\color{codepurple},
    basicstyle=\ttfamily\footnotesize,
    breakatwhitespace=false,         
    breaklines=true,                 
    captionpos=b,                    
    keepspaces=true,                 
    numbers=left,                    
    numbersep=5pt,                  
    showspaces=false,                
    showstringspaces=false,
    showtabs=false,                  
    tabsize=2,
}

\lstdefinelanguage{PyTorch}{%
  language     = Python,
  morekeywords = {ones, T, where, argmax, update, shape, get_min_concept, log_and_keep},
}

\newcommand{\xhdrflat}[1]{\noindent\textbf{#1}}

\newcommand{\myquote}[1]{``\emph{#1}''}

\definecolor{cvprblue}{rgb}{0.21,0.49,0.74}
\usepackage[pagebackref,breaklinks,colorlinks,citecolor=cvprblue]{hyperref}
\usepackage[capitalize]{cleveref}
\crefname{section}{Sec.}{Secs.}
\Crefname{section}{Section}{Sections}
\crefname{table}{Tab.}{Tabs.}
\Crefname{table}{Table}{Tables}

\makeatletter
\DeclareRobustCommand\onedot{\futurelet\@let@token\@onedot}
\def\@onedot{\ifx\@let@token.\else.\null\fi\xspace}

\def\eg{\emph{e.g}\onedot}

 \def\vs{\emph{vs}\onedot}

\makeatother

\title{You Never Know: Quantization Induces Inconsistent Biases in Vision-Language Foundation Models}  
\author{
    {Eric Slyman, Anirudh Kanneganti, Sanghyun Hong, Stefan Lee} \\
    Oregon State University \\
    \texttt{\{slymane, kannegaa, hongsa, leestef\}@oregonstate.edu}
}

\begin{document}
\maketitle

\begin{abstract}
We study the impact of a standard practice in compressing foundation vision-language models--\emph{quantization}--on the models' ability to produce socially-fair outputs. In contrast to prior findings with unimodal models that compression consistently amplifies social biases, our extensive evaluation of four quantization settings across three datasets and three CLIP variants yields a surprising result: while individual models demonstrate bias, we find \emph{no consistent change} in bias magnitude or direction across a population of compressed models due to quantization.
\end{abstract}    

\section{Introduction}
\label{sec:introduction}

Quantization~\citep{gholami2022survey} is a leading practice in compressing deep learning models: it transforms a model's parameter representation from 32-bit floating-point numbers to lower bit-width (e.g., 8-bit or 4-bit integers), thereby reducing memory footprint and inference latency significantly. However, these transformations in number representation can introduce small numerical perturbations to a model's parameter values, potentially leading to undesirable behaviors of a model after quantization~\citep{10.5555/3540261.3540973,9737144,hooker2019what}. In this paper, we study the effects of quantization on the fairness outcomes of foundation vision-language (ViL) models.

\xhdrflat{Related work.} Most work studies compression-induced bias in the unimodal setting, such as vision or language models. \citet{hooker2019what} first noted that the drop in accuracy induced by compressing vision models is concentrated in a few classes which are \myquote{cannibalized} to preserve accuracy in the others. Follow-up work \citep{hooker2020characterising} notes that compression error disparately affects data with low representation in the training distribution that often correlates with socially meaningful features like gender and age. \citet{silva2021towards} similarly find that distilled language models \myquote{almost always exhibit statistically significant bias.} Subsequent works show that
compressing language models amplifies gender bias \citep{renduchintala2021gender,ahn2022why} and that vision model compression has a disparate impact on face classification accuracy \citep{tran2022pruning}, expression recognition \citep{stoychev2022effect}, and other traditional vision tasks \citep{paganini2023prune}.

Recent studies extend these investigations to varied compression techniques across different domains such as facial recognition \citep{lin2022fairgrape}, medical diagnosis \citep{wu2022fairprune}, and multilingual NLP \citep{ogueji2022intriguing, lee2023debiased}. Fairness-aware compression methods analyzed the trade-offs between model fairness, performance, and environmental impact \citep{hessenthaler2022bridging}. \citet{sung2024ecoflap} have even developed a compression technique specifically for ViL models. However, no work to date has studied the fairness impacts of compression for multimodal ViL models, leaving a critical gap on how these techniques affect integrated architectures.

\xhdrflat{Contributions.}
We address this knowledge gap by extensively evaluating quantization effects in multimodal ViL models focusing on fairness outcomes across socially meaningful features like gender, age, and race. Contrary to previous findings in fair compression, our analysis reveales a surprising result: there is no significant evidence for consistent bias amplification across quantized ViL models. While individual models do exhibit biases, the direction and magnitude of these biases were not uniform, suggesting that the impact of compression on fairness may be more nuanced in multimodal contexts. These findings raise questions about the generalizability of compression-induced bias across different model architectures and modalities, indicating a potential need for a more refined understanding of how quantization affects fairness. Additionally, this result may indirectly support recent findings on arbitrariness in fair binary classification \citep{cooper2023arbitrariness}.

\section{Methodology and Experiments}
\begin{figure}[t!]
\centering
\caption{Zero-shot image classification accuracy on ImageNet1K~\citep{deng2009imagenet} and text-image retrieval recall on COCO Captions~\cite{lin2014coco} across varied CLIP versions, training data sources, and quantization methods. Higher ($\uparrow$) is better in all cases. HuggingFace-based quantization methods preserve performance while the PyTorch-based method shows a reduction across metrics.}
\includegraphics[width=0.98\columnwidth]{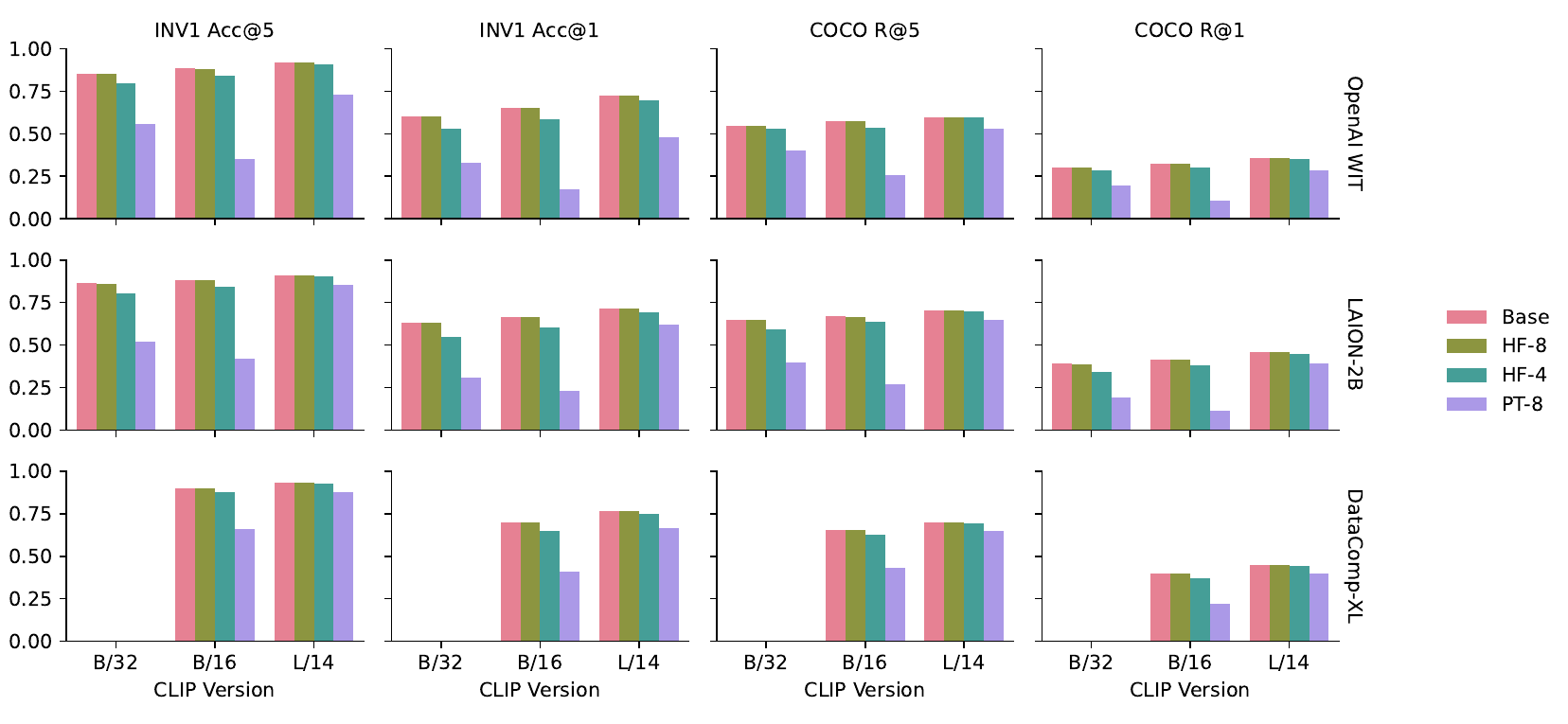}
\label{fig:benchmark}
\vspace{-1.6em}
\end{figure}
\label{sec:experiments}

We examine three common off-the-shelf quantization methods for model compression: 8-bit and 4-bit quantization from the \texttt{bitsandbytes}~\citep{dettmers2022bit} integration into Hugging Face Transformers ~\citep{wolf2020transformers}, and PyTorch's~\citep{paszke2019pytorch} 8-bit dynamic quantization.

\xhdrflat{HuggingFace 8-bit Quantization.}
The 8-bit quantization method, initially introduced by ~\citet{dettmers2024gpt3} in their work on LLM.int8(), represents a significant step towards efficient model compression. This method employs a linear quantization scheme for weight representation, quantizing weights to 8-bit integers while preserving activations in higher precision. It offers up to 50\% reduction in model size compared to FP16 representations and generally yields better performance than lower-bit alternatives, albeit with a smaller compression ratio. A key innovation in this approach is the use of vector-wise quantization, which quantizes vectors (rows or columns) of the weight matrix independently, allowing for better preservation of the weight distribution.

\xhdrflat{HuggingFace 4-bit Quantization.}
Building upon this work, ~\citet{dettmers2024qlora} introduced 4-bit quantization with their QLoRA method, which utilizes the NormalFloat (NF4) data type. This specialized format is optimized for the weight distribution typically observed in neural networks. The 4-bit quantization approach achieves up to 75\% reduction in model size compared to FP16 representations, enabling the loading and inference of larger models on consumer-grade GPUs with limited memory. A crucial aspect of this method is the use of blockwise quantization. In this scheme, the weight matrix is divided into small blocks (typically 64 or 128 elements), and each block is quantized independently. This approach allows for more fine-grained quantization, better preserving the local structure of the weight matrix.

\xhdrflat{Pytorch 8-bit Quantization.}
PyTorch's dynamic quantization ~\citep{paszke2019pytorch} offers a complementary approach that's worth considering. This post-training technique focuses on reducing inference time and memory usage, particularly on CPU architectures. It quantizes weights to 8-bit integers and dynamically quantizes activations during the inference phase, utilizing a dynamic range for activations by calculating scaling factors on-the-fly. This method is well-suited for models with varying input sizes or dynamic computation graphs.

We choose CLIP~\citep{radford2021clip} as a representative foundation alignment model and apply each quantization method above to a variety of model variants across different training data sources. Specifically, we consider the base size vision transformer (ViT)~\citep{dosovitskiy2020image} variants with sequence length 32 (B/32) and 16 (B/16), and the large size ViT variant with sequence length 14 (L/14). For each CLIP variant, we consider a model pretrained on OpenAI WIT~\citep{radford2021clip}, LAION-2B~\citep{schuhmann2022laion}, and DataComp-XL~\citep{gadre2024datacomp}. We find the weights available online for the B/32 varient trained on DataComp-XL to be corrupted, resulting in an evaluation over eight distinct models with three quantization methods totalling 32 scenarios.

\begin{table}[t!]
\centering
\caption{P-value and 95\% confidence intervals for paired t-tests on FACET disparity pre- and post-quantization difference in means. Cells in \textcolor{Dandelion}{yellow}/\textcolor{ForestGreen}{green} are significant at \textcolor{Dandelion}{90\%}/\textcolor{ForestGreen}{95\%} confidence.}
\label{tab:facet}
\resizebox{0.9\linewidth}{!}{%
\begin{tabular}{llr@{\ \ }r@{,\ \ }rr@{\ \ }r@{,\ \ }rr@{\ \ }r@{,\ \ }rr@{\ \ }r@{,\ \ }r}
\toprule
\textbf{Quant Method} &
  \textbf{Demographic} &
  \multicolumn{3}{l}{\textbf{Min Disp@1}} &
  \multicolumn{3}{l}{\textbf{Max Disp@1}} &
  \multicolumn{3}{l}{\textbf{Min Disp@5}} &
  \multicolumn{3}{l}{\textbf{Max Disp@5}} \\
 &
   &
  \multicolumn{1}{l}{\small $p$} &
  \multicolumn{2}{l}{\small $95\%CI$} &
  \multicolumn{1}{l}{\small $p$} &
  \multicolumn{2}{l}{\small $95\%CI$} &
  \multicolumn{1}{l}{\small $p$} &
  \multicolumn{2}{l}{\small $95\%CI$} &
  \multicolumn{1}{l}{\small $p$} &
  \multicolumn{2}{l}{\small $95\%CI$} \\ \midrule
\textbf{HuggingFace} &
  Gender (MF) &
  .862 &
  -.01 &
  .01 &
  \cellcolor[HTML]{FCE8B2}.082 &
  \cellcolor[HTML]{FCE8B2}-.02 &
  \cellcolor[HTML]{FCE8B2}.00 &
  .246 &
  -.01 &
  .00 &
  .356 &
  -.05 &
  .11 \\
4-Bit &
  Skin Tone (LD) &
  .307 &
  -.02 &
  .04 &
  \cellcolor[HTML]{B7E1CD}.035 &
  \cellcolor[HTML]{B7E1CD}.01 &
  \cellcolor[HTML]{B7E1CD}.10 &
  .164 &
  -.01 &
  .03 &
  .210 &
  -.04 &
  .12 \\
 &
  Age (MY) &
  .774 &
  -.01 &
  .01 &
  .425 &
  -.04 &
  .08 &
  .133 &
  -.01 &
  .00 &
  .587 &
  -.05 &
  .09 \\
 &
  Age (MO) &
  .262 &
  -.01 &
  .00 &
  .141 &
  -.13 &
  .03 &
  .501 &
  .00 &
  .00 &
  .388 &
  -.04 &
  .09 \\ \midrule
\textbf{HuggingFace} &
  Gender (MF) &
  .310 &
  .00 &
  .01 &
  .945 &
  -.01 &
  .01 &
  .999 &
  .00 &
  .00 &
  .502 &
  -.03 &
  .05 \\
8-Bit &
  Skin Tone (LD) &
  .700 &
  -.01 &
  .02 &
  .843 &
  -.05 &
  .06 &
  .817 &
  -.01 &
  .01 &
  .447 &
  -.05 &
  .03 \\
 &
  Age (MY) &
  .765 &
  .00 &
  .00 &
  .868 &
  -.03 &
  .03 &
  .456 &
  .00 &
  .00 &
  .245 &
  -.01 &
  .02 \\
 &
  Age (MO) &
  .396 &
  -.01 &
  .00 &
  .269 &
  -.03 &
  .01 &
  .340 &
  .00 &
  .01 &
  .444 &
  -.06 &
  .12 \\ \midrule
\textbf{PyTorch} &
  Gender (MF) &
  .115 &
  -.01 &
  .03 &
  .192 &
  -.05 &
  .16 &
  .278 &
  -.03 &
  .01 &
  \cellcolor[HTML]{FCE8B2}.076 &
  \cellcolor[HTML]{FCE8B2}-.02 &
  \cellcolor[HTML]{FCE8B2}.20 \\
8-Bit &
  Skin Tone (LD) &
  .205 &
  -.01 &
  .04 &
  .247 &
  -.07 &
  .17 &
  .399 &
  -.04 &
  .02 &
  .158 &
  -.08 &
  .02 \\
 &
  Age (MY) &
  .117 &
  -.01 &
  .03 &
  .850 &
  -.17 &
  .15 &
  \cellcolor[HTML]{FCE8B2}.095 &
  \cellcolor[HTML]{FCE8B2}-.02 &
  \cellcolor[HTML]{FCE8B2}.00 &
  .940 &
  -.23 &
  .22 \\
 &
  Age (MO) &
  \cellcolor[HTML]{B7E1CD}.045 &
  \cellcolor[HTML]{B7E1CD}.00 &
  \cellcolor[HTML]{B7E1CD}.04 &
  .462 &
  -.14 &
  .24 &
  .159 &
  -.02 &
  .01 &
  .348 &
  -.26 &
  .13 \\ \bottomrule
\end{tabular}%
}
\end{table}
\subsection{Evaluation Datasets and Metrics}
We evaluate across three benchmarks to validate if quantized models are both \emph{accurate} and \emph{fair}.

\xhdrflat{Zero-Shot Classification and Retrieval.}
We evaluate the accuracy of each model and its quantized variants across two common benchmark tasks for foundation vision-language alignment models: zero-shot image classification on ImageNet~\citep{deng2009imagenet} and text-based image retrieval on COCO ~\citep{lin2014coco}. Quantized variants \emph{should} have similar accuracy to the original model.

\xhdrflat{Fair Zero-Shot Classification.}
The FACET~\citep{gustafson2023facet} dataset contains expert image annotations for 52 person related classes, age, skin tone, and gender presentation. We perform zero-shot classification over the person-classes by constructing a text prompt for each class and predicting the class used to construct the prompt with highest similarity to the image. Following \citet{gustafson2023facet}, we measure the disparity between pairs of values within a sensitive group (\eg $\{light,\ dark\} \in skintone$) as the difference in recall between true-positive instances of a person-class for each value. A large magnitude disparity indicates that a model better predicts positive instances for one member of the group, while a disparity of zero indicates group \emph{equality of opportunity}. We study the the maximum and minimum disparity measured across person-classes.
We utilize the same sensitive groups as \citet{slyman2024fairdedup}, evaluating perceived gender expression by masculine \vs feminine presentation, lighter (1-4MST\footnote{Monk Skin Tone scale \cite{monk2019skin}}) \vs darker (6-10MST) skin tone, and middle \vs younger/older age for all person-classes which have at least 25 samples in both subgroups.

\xhdrflat{Fair Image Retrieval.}
FairFace~\citep{karkkainen2021fairface} annotates cropped faces with perceived race, age, and gender. Following \citep{seth2023dear}, we assess how much the top-$k$ results of an image-text query differ across sensitive attribute values in the validation set using MaxSkew@k~\citep{geyik2019fairness}. For a given top-$k$ image set $\tau_r^k$ from query $r$, let $P_{\tau_r^k,a_i}{\in}[0,1]$ be the actual proportion of images with a particular value $a_i{\in}A$ of sensitive attribute $A$, and $P_{r,a_i}{\in}[0,1]$ be the desired proportion estimated from true rates in the full dataset. Then the skew for $a_i$ is: 
\begin{equation}
Skew_{a_i}@k(\tau_r) = \ln\left(\frac{P_{\tau_r^k,a_i}}{P_{r,a_i}}\right) 
\label{eq:skew}
\end{equation}%
Skew@k is specific to a single value of a sensitive attribute. To provide a more comprehensive view, we report the most (least) skewed attribute value as MaxSkew@k (MinSkew@k). MaxSkew@k indicates the \myquote{largest unfair advantage}~\citep{geyik2019fairness} given to images with a particular attribute value in the top-$k$ results while MinSkew@k captures the%
\begin{wraptable}{r}{0.6\linewidth}
\vspace{-1em}
\centering
\caption{P-value and confidence interval for paired t-tests on FairFace skew pre- and post-quantization means. Cells in \textcolor{Dandelion}{yellow}/\textcolor{ForestGreen}{green} are significant at \textcolor{Dandelion}{90\%}/\textcolor{ForestGreen}{95\%} confidence.}
\label{tab:fairface}
\resizebox{\linewidth}{!}{%
\begin{tabular}{llr@{\ \ }r@{,\ \ }rr@{\ \ }r@{,\ \ }r}
\toprule
\textbf{Quant Method} &
  \textbf{Demographic} &
  \multicolumn{3}{l}{\textbf{MinSkew}} &
  \multicolumn{3}{l}{\textbf{MaxSkew}} \\
 &
   &
  \multicolumn{1}{l}{\small $p$} &
  \multicolumn{2}{l}{\small $95\%CI$} &
  \multicolumn{1}{l}{\small $p$} &
  \multicolumn{2}{l}{\small $@95\%CI$} \\ \midrule
\textbf{HuggingFace} &
  Gender &
  \cellcolor[HTML]{FCE8B2}.066 &
  \cellcolor[HTML]{FCE8B2}-.19 &
  \cellcolor[HTML]{FCE8B2}.01 &
  \cellcolor[HTML]{B7E1CD}.039 &
  \cellcolor[HTML]{B7E1CD}-.07 &
  \cellcolor[HTML]{B7E1CD}.01 \\
4-Bit &
  Race &
  .225 &
  -.04 &
  .12 &
  \cellcolor[HTML]{FCE8B2}.073 &
  \cellcolor[HTML]{FCE8B2}-.01 &
  \cellcolor[HTML]{FCE8B2}.06 \\
 &
  Age &
  .557 &
  -.10 &
  .06 &
  .800 &
  -.04 &
  .03 \\ \midrule
\textbf{HuggingFace} &
  Gender &
  .632 &
  -.01 &
  .02 &
  .655 &
  -.01 &
  .01 \\
8-Bit &
  Race &
  .717 &
  -.01 &
  .01 &
  \cellcolor[HTML]{FCE8B2}.060 &
  \cellcolor[HTML]{FCE8B2}.00 &
  \cellcolor[HTML]{FCE8B2}.01 \\
 &
  Age &
  .153 &
  -.02 &
  .01 &
  .157 &
  -.01 &
  .01 \\ \midrule
\textbf{PyTorch} &
  Gender &
  .560 &
  -.08 &
  .13 &
  .535 &
  -.03 &
  .05 \\
8-Bit &
  Race &
  .938 &
  -.13 &
  .12 &
  .555 &
  -.03 &
  .06 \\
 &
  Age &
  \cellcolor[HTML]{FCE8B2}.077 &
  \cellcolor[HTML]{FCE8B2}-.02 &
  \cellcolor[HTML]{FCE8B2}.26 &
  .124 &
  -.01 &
  .07 \\ \bottomrule
\end{tabular}%
}
\end{wraptable}
\myquote{worst disadvantage in representation} for a subgroup.
With the condition that the desired proportion of images matches the true distribution of those images in the dataset, an optimal MaxSkew@k of $0$ can be shown to achieve the fairness criteria of \emph{demographic parity.} Following \citet{berg2022prompt}, we report average MaxSkew@1000 across 240 (un)favorable captions orthogonal to images in the dataset, matching test attributes and prompts for race, age, and gender. Similar to \citet{slyman2024fairdedup}, we reduce noise by binning age into: \emph{younger} (0-19), \emph{middle} (20-49), and \emph{older} (50-70+) subgroups.

\section{Empirical Evaluation}
\label{sec:results}

\xhdrflat{Accuracy.} As shown in Fig.~\ref{fig:benchmark}, the selected quantization methods generally preserve accuracy across different models and tasks. This result indicates that the methods chosen for our study are effective in terms of preserving baseline performance. A method which does not preserve performance (\eg resulting in random predictions) may otherwise trivially fulfil many common fairness criteria.

\xhdrflat{Fairness.} We consider fairness evaluations as paired pre-post/quantization measurements and assess the significance of difference between the two measures with a paired t-test. The results for fairness are mixed. Table~\ref{tab:facet} presents the outcomes for FACET, where we observe inconsistent equality of opportunity outcomes in zero-shot image classification. The disparities across different demographic groups vary, with some quantization methods leading to minor but somewhat statistically significant changes. As shown in Table~\ref{tab:fairface}, FairFace demonstrates similar inconsistent skew outcomes in image retrieval. We note that these observations are without correcting for multiple testing, and that the minor significant results observed here are likely to disappear under most correction methods. 

\xhdrflat{Limitations.}
Our evaluations make several generalizability limiting assumptions. Specifically, we study only CLIP models under quantization as the compression method. It would be a compelling line of future work to understand fairness outcomes when applying more advanced compression methods (\eg, pruning or distillation), alternative alignment model architectures, or more advanced ViL models (\eg BLIP~\citep{li2023blip}) which can perform VQA and image captioning tasks.

\section{Conclusion}
\label{sec:conclusion}
Our study reveals that the impact of quantization on bias in multimodal ViL models is neither consistent nor uniform across different models, methods, and datasets. The direction and magnitude of bias introduced by quantization varied, indicating that its effects on fairness are complex and context-dependent. These findings challenge the assumption that quantization consistently influences bias across settings, highlighting the need for a more nuanced understanding of how compression techniques impact fairness across diverse model architectures and applications.

\xhdrflat{Acknowledgments.}
Anirudh and Sanghyun are supported by the Samsung Global Research Outreach (GRO) Program 2024.
Any opinions, findings, and conclusions or recommendations expressed here are those of the authors 
and do not necessarily reflect the views of the funding agencies.

{
    \small
    \bibliographystyle{style/ieeenat_fullname}
    \bibliography{main}
}

\end{document}